\documentclass{article} % For LaTeX2e
\usepackage{iclr2026_conference,times}
\usepackage{kotex}
\usepackage{algorithm}
\usepackage{algpseudocode}
\usepackage{graphicx}
\usepackage{subcaption}
\usepackage{booktabs}

\usepackage{amsmath,amsfonts,bm}

\def\eqref#1{equation~\ref{#1}}
\def\1{\bm{1}}

\DeclareMathAlphabet{\mathsfit}{\encodingdefault}{\sfdefault}{m}{sl}
\SetMathAlphabet{\mathsfit}{bold}{\encodingdefault}{\sfdefault}{bx}{n}

\usepackage{hyperref}
\usepackage{url}

\iclrfinalcopy

\title{Inference-Time Scaling in Diffusion Models through Iterative Partial Refinement}

\author{
\hspace{-3.0em}
\begin{tabular}[t]{c}
\bf Taegu Kang \\
\normalfont KAIST \\
\normalfont\texttt{rivertae9@kaist.ac.kr}
\end{tabular}\;
\begin{tabular}[t]{c}
\bf Jaesik Yoon \\
\normalfont KAIST \& SAP \\
\normalfont\texttt{jaesik.yoon@kaist.ac.kr}
\end{tabular}\;
\begin{tabular}[t]{c}
\bf Sungjin Ahn \\
\normalfont KAIST \& NYU \\
\normalfont\texttt{sungjin.ahn@kaist.ac.kr}
\end{tabular}
}

\begin{document}

\maketitle

\begin{abstract}
Inference-time scaling has emerged as a major approach for improving reasoning capabilities, and has been increasingly applied to diffusion models. However, existing inference-time scaling methods for diffusion models typically rely on external verifiers or reward models to rank and select samples, limiting their scalability to settings where such evaluators are available and reliable. Moreover, while recent diffusion models perform \emph{sequential} inference with region-wise, mixed-noise conditioning, inference-time scaling tailored to this setting remains relatively underexplored. We propose \textbf{Iterative Partial Refinement (IPR)}, an inference-time scaling method for sequential diffusion that requires no external verifier. Starting from an already-generated sample, IPR re-noises a subset of regions and regenerates them conditioned on the remaining regions, enabling the model to revise earlier decisions under a richer context than was available during the initial generation. This iterative partial refinement produces more globally consistent samples without external verification. On reasoning tasks requiring global constraint satisfaction, IPR consistently improves performance: on \textbf{MNIST Sudoku}, the valid solution rate increases from \textbf{55.8\%} to \textbf{75.0\%}. These results show that iterative partial refinement alone can serve as an effective inference-time scaling strategy for diffusion models in sequential, mixed-noise settings. Code is available at: \url{https://github.com/ahn-ml/IPR}

\end{abstract}

\section{Introduction}
\label{sec:intro}

Diffusion models are strong generators~\citep{ho2020denoising, song2020denoising}, and a growing line of work studies \emph{inference-time scaling}---improving generations by spending additional compute at test time~\citep{singhal2025general, li2024derivative, kim2025test, guo2025training, lee2025adaptive, zhang2025inference, he2025scaling}.
Most existing approaches perform inference-time scaling by generating multiple candidates or trajectories and then ranking, selecting, or resampling them using an external reward model, verifier, or task-specific scoring function.
While effective, this reliance restricts applicability to settings where such evaluators are unavailable or unreliable.

Recently, VFScale \citep{zhang2025vfscale} takes an important step toward removing external evaluators by using the diffusion model's intrinsic energy as an internal scoring signal and scaling test-time compute through search over denoising trajectories.
However, verifier-free scaling in this form is primarily developed for standard DDPM-style diffusion, where all regions are updated at the same noise level at each step.

Many realistic generation problems are inherently compositional: different parts of the output must agree with each other, so decisions made in one region constrain what is plausible in others.
Such cross-region dependencies are often the norm rather than an exception, and they call for inference mechanisms that can exploit \emph{asymmetric} conditioning between regions.
% Standard diffusion, with its uniform-noise updates, offers limited structure for such conditional generation across regions.
Standard diffusion, with its uniform-noise updates, offers limited structure for exploiting such dependencies.

To address these limitations, recent works have moved toward modeling noise levels independently across different regions, decoupling the noise schedule from a single global timestep. This approach, which we refer to as \textbf{sequential diffusion}~\citep{chen2024diffusion, wewer2025spatial, wu2023ar, zhang2024tedi, li2024autoregressive}, allows different regions to evolve at varying noise levels so that less-noisy regions provide stable context for generating noisier ones. By training on diverse region-wise noise configurations, these models learn mixed-noise conditional denoising behavior, making cross-region dependencies more explicit and enabling more structured conditional generation.

Despite this promise, inference-time scaling for sequential diffusion has received relatively less attention than its standard-diffusion counterpart.
In particular, while sequential diffusion provides a natural mechanism for conditioning across regions, it remains unclear how to allocate additional inference-time compute to improve samples \emph{without relying on external verifiers or rewards}.
This gap motivates our work.

We propose \textbf{Iterative Partial Refinement (IPR)}, a simple inference-time scaling method for sequential diffusion models trained on arbitrary noise levels.
Starting from an initial sample, IPR repeatedly selects a subset of regions, re-noises them, and regenerates them conditioned on the remaining regions.
This operation allows inference to revisit earlier decisions and correct inconsistencies that would otherwise persist.
IPR requires no additional training, no guidance, and no external verifier; it simply reuses the model's learned conditional distribution for self-improvement.

We evaluate IPR on globally constrained generation benchmarks: \textbf{MNIST Sudoku}, \textbf{Even Pixels}, and \textbf{Counting Polygons}.
Across tasks, increasing the number of IPR iterations consistently improves constraint satisfaction and sample quality; on MNIST Sudoku HARD, IPR increases the valid solution rate from 55.8\% to 75.0\%.
Overall, these results show that enabling revision during inference provides a simple way to scale sequential diffusion at inference time, improving global consistency without modifying the underlying model parameters.

Our contributions are as follows:
\begin{itemize}
    \item \textbf{Inference-Time Refinement Strategy:} We introduce IPR, a method that revises generated samples by iteratively re-noising and regenerating sub-regions. By targeting the inference phase, we enable correction of global inconsistencies without permanent parameter updates.
    \item \textbf{Self-Improvement via Intrinsic Capability:} We demonstrate that the model can correct its own errors by leveraging its native mixed-noise conditional denoising capability. This approach achieves self-improvement using only the pre-trained model, avoiding the need for external reward models or verifiers.
    \item \textbf{Validation on Structured Generation:} We empirically validate IPR on three globally constrained benchmarks (MNIST Sudoku, Even Pixels, Counting Polygons), showing that simple iterative refinement significantly enhances constraint satisfaction rates and improves the quality of generations.
\end{itemize}

\section{Preliminary}
\label{sec:prelim}

\subsection{Diffusion Generative Models}
Let $\mathbf{x}_0 \in \mathbb{R}^D$ be a data sample drawn from $p_0(\mathbf{x}_0)$. Diffusion models~\citep{ho2020denoising, song2020denoising} and flow matching~\citep{lipman2022flow} both construct a stochastic process $\{\mathbf{x}_t\}_{t=0}^{T}$ that continuously transforms data into noise. We refer to the continuous index $t\in[0,1]$ as the \emph{noise level}, where $t{=}0$ corresponds to clean data and $t{=}1$ to pure Gaussian noise. Given a noise schedule $(\alpha_t, \sigma_t)$, a pair of monotonic functions satisfying the boundary conditions $\alpha_0{=}1,\;\sigma_0{=}0$ and $\alpha_1{=}0,\;\sigma_1{=}1$ (i.e., the signal-to-noise ratio $\alpha_t/\sigma_t$ decreases monotonically from $\infty$ to $0$), the noisy variable at noise level $t$ is defined as the interpolation
\begin{equation}
    \mathbf{x}_t = \alpha_t\,\mathbf{x}_0 + \sigma_t\,\boldsymbol{\epsilon}, \quad \boldsymbol{\epsilon}\sim\mathcal{N}(\mathbf{0},\mathbf{I}).
\end{equation}
A neural network $\boldsymbol{\epsilon}_\theta$ is trained to recover the noise component via the denoising objective
\begin{equation}
    \mathcal{L}(\theta)
    = \mathbb{E}_{t,\,\mathbf{x}_0,\,\boldsymbol{\epsilon}}\!\left[\left\|\boldsymbol{\epsilon}_\theta(\mathbf{x}_t, t) - \boldsymbol{\epsilon}\right\|^2\right].
\end{equation}
To generate samples, one draws $\mathbf{x}_1\sim\mathcal{N}(\mathbf{0},\mathbf{I})$ and iteratively applies the learned reverse process $p_\theta(\mathbf{x}_{t-\Delta t}\mid\mathbf{x}_t)$, using either deterministic updates~\citep{song2020denoising} or stochastic sampling~\citep{ho2020denoising}, progressing from $t{=}1$ down to $t{=}0$. We refer to this setup, where all regions share the same noise level and are denoised in parallel, as \emph{standard diffusion} throughout the paper.

\subsection{Sequential Diffusion Models}
Unlike standard diffusion, where all regions share a single noise level, \emph{sequential diffusion}~\citep{chen2024diffusion, wewer2025spatial, wu2023ar, zhang2024tedi, li2024autoregressive} assigns each region its own noise level. Regions denoised earlier provide partially clean context for the remaining, noisier regions, making cross-region dependencies explicit.

Formally, let an output $\mathbf{x}$ be partitioned into $N$ regions $\{x_1,\dots,x_N\}$ (for images, each region is a patch). Let $x_i^{t}$ denote region $i$ at noise level $t$, and $\mathbf{t}=(t_1,\dots,t_N)$ be a vector of per-region noise levels. Given a source configuration $\mathbf{t}'$, the model predicts the denoised output at a target configuration $\mathbf{t}$ ($t_i \leq t'_i$):
\begin{equation}
p_\theta\!\left(x_1^{t_1},\dots,x_N^{t_N}\,\middle|\,x_1^{t'_1},\dots,x_N^{t'_N}\right).
\end{equation}
% Training samples diverse pairs $(\mathbf{t}',\mathbf{t})$ so the model learns to denoise regions conditioned on the rest. The choice of training noise distribution determines the model's capability: one can enforce a fixed generation order, or sample arbitrary mixed noise levels to enable the model to capture complex, order-agnostic dependencies.
Training involves diverse pairs $(\mathbf{t}',\mathbf{t})$ so that the model learns to denoise regions conditioned on the rest. The choice of training noise distribution determines the model's capability: one can enforce a fixed generation order, or sample arbitrary mixed noise levels to enable the model to capture complex, order-agnostic dependencies.

\textbf{Spatial Reasoning Models (SRMs)}~\citep{wewer2025spatial} instantiate this framework for images, training on patch-wise mixed-noise configurations. At inference, SRMs select which patch to denoise next based on prediction uncertainty, denoising confident patches first to build reliable context, and we adopt SRMs as our sequential diffusion backbone.
\section{Iterative Partial Refinement}
\label{sec:method}
% \begin{figure}[t]
%     \centering
%     \includegraphics[width=1.0\linewidth]{figures/figure1.png}
%     \caption{\textbf{Progressive refinement on MNIST Sudoku with IPR.}
% From left to right, Iterative Partial Refinement (IPR) progressively corrects erroneous cells (highlighted in red), resolving global constraint violations and converging to a valid Sudoku solution.}
%     \label{fig:sudoku_refine}
% \end{figure}

\begin{figure}[t]
    \centering
    \includegraphics[width=1.0\linewidth]{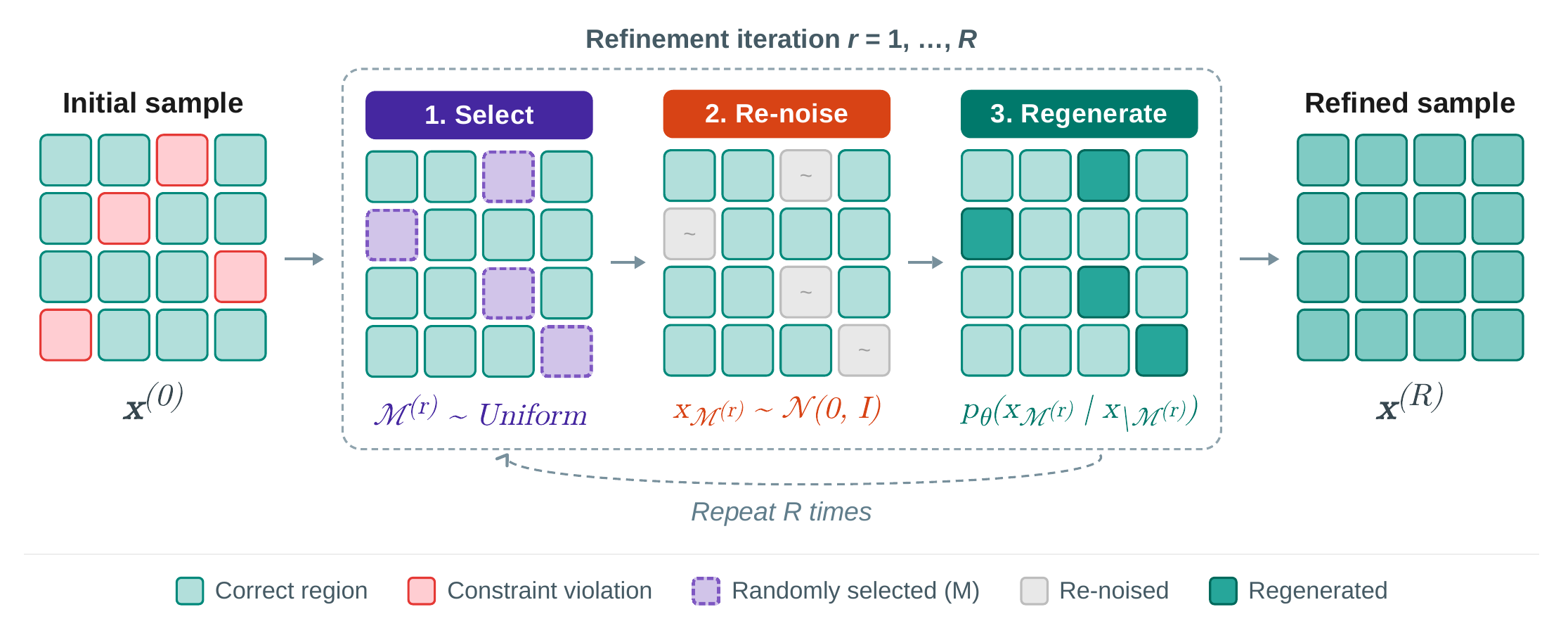}
    \caption{\textbf{Overview of Iterative Partial Refinement (IPR).}
    Starting from an initial sample $\mathbf{x}^{(0)}$, IPR iteratively refines the output by repeating three steps: (1) randomly selecting a subset $\mathcal{M}^{(r)}$ of regions, (2) replacing them with random noise, and (3) regenerating them conditioned on the remaining regions via the learned conditional distribution $p_\theta(\mathbf{x}_{\mathcal{M}^{(r)}} \mid \mathbf{x}_{\setminus\mathcal{M}^{(r)}})$. After $R$ iterations, the refined sample $\mathbf{x}^{(R)}$ is returned.}
    \label{fig:ipr_method}
\end{figure}

Sequential diffusion follows a \textbf{one-pass generation trajectory}. Early in sampling, the context is formed under high noise and limited information, so stochastic updates may introduce small inconsistencies. Because later steps condition on this context without revisiting it, such inconsistencies can persist and propagate to the rest of the sequence. To address this limitation, we propose \textbf{Iterative Partial Refinement (IPR)}, an inference-time mechanism for sequential diffusion models trained on arbitrary noise levels. Starting from a generated sample, IPR repeatedly selects a subset of regions, adds noise to that subset, and regenerates it conditioned on the remaining regions. Figure~\ref{fig:ipr_method} illustrates the overall procedure of IPR.

\paragraph{Setup.}
We denote the output as $\mathbf{x} = (x_1, \dots, x_N)$, where each $x_i$ corresponds to a region. Sequential diffusion allows regions to reside at different noise levels, and the model $p_\theta$ is trained to perform conditional denoising under mixed noise levels.

\paragraph{Procedure.}
Given an initial sample $\mathbf{x}^{(0)}$ from standard sequential inference, IPR performs $R$ refinement iterations. At each iteration $r = 1, \dots, R$:
\begin{enumerate}
    \item \textbf{Select:} Sample a random subset $\mathcal{M}^{(r)} \subset \{1, \dots, N\}$ with $|\mathcal{M}^{(r)}| = \lfloor\alpha N\rfloor$, where $\alpha$ is the resampling ratio. Regions provided as fixed input conditions are excluded from the sampling pool.
    \item \textbf{Re-noise:} Replace regions in $\mathcal{M}^{(r)}$ with random noise, i.e., sample $\mathbf{x}^{(r)}_{\mathcal{M}^{(r)}} \sim \mathcal{N}(0, I)$.
    \item \textbf{Regenerate:} Conditionally regenerate the selected regions given the remaining context (where $\setminus \mathcal{M}^{(r)}$ denotes the complement $\{1, \dots, N\} \setminus \mathcal{M}^{(r)}$):
    \begin{align}
        \mathbf{x}^{(r)}_{\mathcal{M}^{(r)}} &\sim p_\theta\bigl(\mathbf{x}_{\mathcal{M}^{(r)}} \mid \mathbf{x}^{(r-1)}_{\setminus \mathcal{M}^{(r)}}\bigr), \\
        \mathbf{x}^{(r)}_{\setminus \mathcal{M}^{(r)}} &= \mathbf{x}^{(r-1)}_{\setminus \mathcal{M}^{(r)}}.
    \end{align}
\end{enumerate}
After $R$ iterations, the final sample $\mathbf{x}^{(R)}$ is returned. If standard sequential inference requires $T$ total denoising steps to generate all $N$ regions, IPR adds $O(\alpha R T)$ steps, corresponding to regenerating $\alpha N$ regions per iteration over $R$ rounds. Algorithm~\ref{alg:ipr} summarizes the procedure.

\paragraph{Intuition.}
Repeating this process gives inference multiple opportunities to revise earlier mistakes. Re-noising part of the current sample undoes fixed decisions that were made under limited context, and lets the model regenerate that part given the now more complete surrounding regions. Over iterations, this can rebuild contexts closer to those seen during training, making it easier to correct inconsistencies and improve global consistency. In this way, reusing the model's own learned distribution for iterative refinement serves as an effective inference-time scaling strategy—achieving gains without requiring external verifiers or task-specific heuristics.

% \begin{algorithm}[t]
% \caption{Iterative Partial Refinement (IPR)}
% \label{alg:ipr}
% \begin{algorithmic}[1]
% \Require Trained sequential diffusion model $p_\theta$, iterations $R$, resampling ratio $\alpha$, fixed condition set $\mathcal{C}$
% \State Initialize $\mathbf{x}^{(0)}$ via sequential diffusion inference
% \For{$r = 1$ to $R$}
% \State Sample $\mathcal{M}^{(r)} \subset \{1,\dots,N\} \setminus \mathcal{C}$ uniformly at random, $|\mathcal{M}^{(r)}| = \lfloor\alpha (N - |\mathcal{C}|)\rfloor$
% \State Replace $\mathbf{x}^{(r-1)}_{\mathcal{M}^{(r)}}$ with random noise $\sim \mathcal{N}(0, I)$
% \State $\mathbf{x}^{(r)}_{\mathcal{M}^{(r)}} \sim p_\theta\bigl(\mathbf{x}_{\mathcal{M}^{(r)}} \mid \mathbf{x}^{(r-1)}_{\setminus \mathcal{M}^{(r)}}\bigr)$
% \State $\mathbf{x}^{(r)}_{\setminus \mathcal{M}^{(r)}} \leftarrow \mathbf{x}^{(r-1)}_{\setminus \mathcal{M}^{(r)}}$
% \EndFor
% \State \Return $\mathbf{x}^{(R)}$
% \end{algorithmic}
% \end{algorithm}

\begin{algorithm}[t]
\caption{Iterative Partial Refinement (IPR)}
\label{alg:ipr}
\begin{algorithmic}[1]
\Require Trained sequential diffusion model $p_\theta$, iterations $R$, resampling ratio $\alpha$, condition set $\mathcal{C} \subseteq \{1,\dots,N\}$ (default $\emptyset$)
\State Initialize $\mathbf{x}^{(0)}$ via sequential diffusion inference
\For{$r = 1$ to $R$}
\State Sample $\mathcal{M}^{(r)} \subset \{1,\dots,N\} \setminus \mathcal{C}$ uniformly at random, $|\mathcal{M}^{(r)}| = \lfloor\alpha (N - |\mathcal{C}|)\rfloor$
\State Replace $\mathbf{x}^{(r-1)}_{\mathcal{M}^{(r)}}$ with random noise $\sim \mathcal{N}(0, I)$
\State $\mathbf{x}^{(r)}_{\mathcal{M}^{(r)}} \sim p_\theta\bigl(\mathbf{x}_{\mathcal{M}^{(r)}} \mid \mathbf{x}^{(r-1)}_{\setminus \mathcal{M}^{(r)}}\bigr)$
\State $\mathbf{x}^{(r)}_{\setminus \mathcal{M}^{(r)}} \leftarrow \mathbf{x}^{(r-1)}_{\setminus \mathcal{M}^{(r)}}$
\EndFor
\State \Return $\mathbf{x}^{(R)}$
\end{algorithmic}
\end{algorithm}

\section{Experiments}
\label{sec:experiments}

\subsection{Experimental Setup}

\begin{figure}[t]
    \centering
    \includegraphics[width=1.0\linewidth]{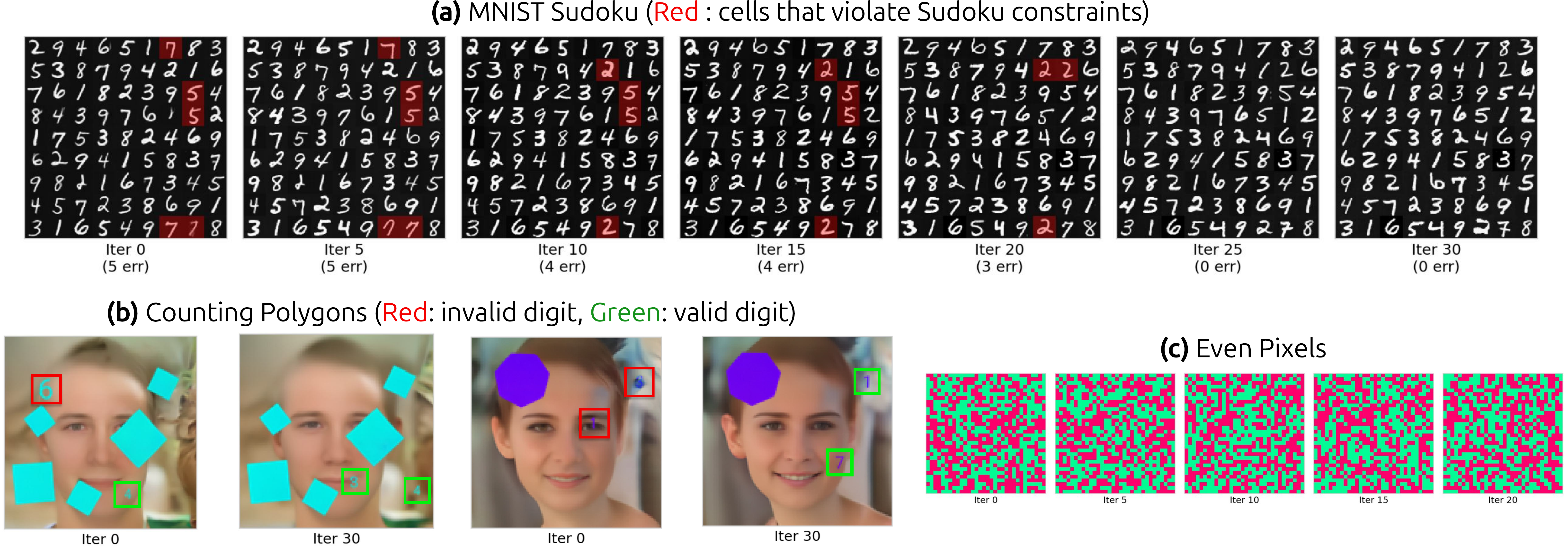}
    \caption{\textbf{Qualitative results across three benchmarks.}
    \textbf{(a)} On MNIST Sudoku, IPR progressively corrects constraint-violating cells (red background), converging to a valid solution.
    \textbf{(b)} On Counting Polygons, IPR refines the generated digits to match the polygon count and vertex type (red: invalid, green: valid).
    \textbf{(c)} On Even Pixels, color regions become more uniform and balanced over iterations.}
    \label{fig:qualitative}
\end{figure}

\paragraph{Backbone and Benchmarks.}
We use \textbf{Spatial Reasoning Models (SRMs)}~\citep{wewer2025spatial} as the backbone, a general framework for sequential diffusion that supports patch-wise noise configurations. To ensure fair comparison, we adopt the same constrained generation benchmarks used in the original SRMs work: \emph{MNIST Sudoku}, \emph{Counting Polygons}, and \emph{Even Pixels}. Each benchmark requires satisfying different types of global constraints. Qualitative examples across all three benchmarks are shown in Figure~\ref{fig:qualitative}, with additional results in Appendix~\ref{qual_appendix}.

\paragraph{Baseline and Evaluation.}
We compare IPR against the standard SRMs inference (without refinement) as the baseline. Common inference-time scaling strategies such as best-of-$n$ sampling or pass@$k$ require external verifiers to select or rank candidates; since our goal is to evaluate whether IPR can improve generation quality by better utilizing the learned distribution alone, we do not use such verifiers. All experiments use the same pretrained SRMs models without additional training.

\paragraph{Hyperparameters.}
Unless otherwise specified, we fix the resampling ratio $\alpha = 0.25$ and measure performance as a function of refinement iterations $R$. We use $R \le 50$ in practice and do not employ early stopping. Detailed hyperparameters for each task are provided in Appendix~\ref{app:hyperparameters}.

\subsection{MNIST Sudoku}
\textbf{MNIST Sudoku} follows standard Sudoku rules: each row, column, and $3 \times 3$ subgrid must contain the digits 1--9 without repetition. The key difference is that each cell is represented as a $28 \times 28$ MNIST digit image, requiring the model to generate visually coherent digits while satisfying combinatorial constraints. We use SRMs trained on 81 patches (one per cell) to generate the full grid (see Appendix~\ref{app:mnist_sudoku}).

\begin{figure}[t]
    \centering
    \begin{subfigure}[t]{0.48\textwidth}
        \centering
        \includegraphics[width=0.85\linewidth]{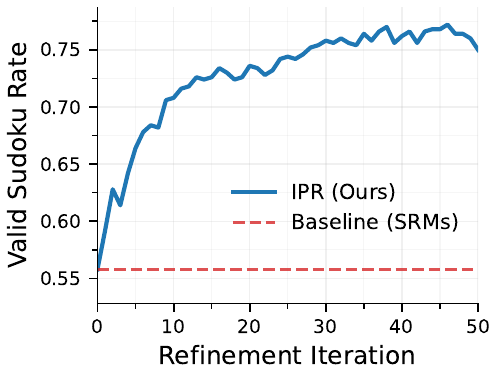}
        \caption{Refinement on HARD Sudoku}
        \label{fig:sudoku_acc}
    \end{subfigure}
    \hfill
    \begin{subfigure}[t]{0.48\textwidth}
        \centering
        \includegraphics[width=0.85\linewidth]{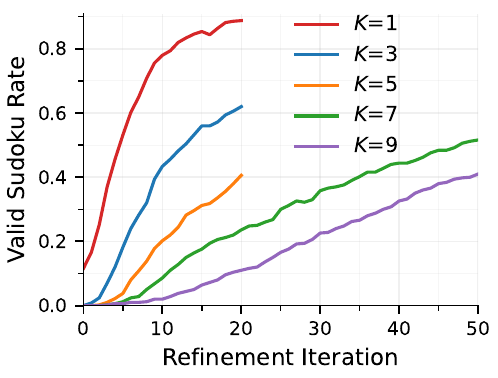}
        \caption{Refinement on $K$-Corrupted Sudoku}
        \label{fig:corrupt_sudoku}
    \end{subfigure}
    \caption{\textbf{IPR on MNIST Sudoku.} (a) Valid Sudoku rate improves consistently with IPR iterations on the HARD setting. (b) Recovery rate from corrupted grids with $K$ initially swapped cell pairs, showing robustness even under severe corruption.}
    \label{fig:sudoku_results}
\end{figure}

\paragraph{HARD Setting.}
We evaluate on the HARD setting, where only 0--26 cells are given as clues. During IPR, clue cells are fixed and excluded from resampling. As shown in Figure~\ref{fig:sudoku_acc}, the baseline SRMs achieves a 55.8\% valid Sudoku rate, while IPR progressively improves this to over 75\%. This demonstrates that iteratively re-noising and regenerating parts of the sample allows the model to correct inconsistencies and improve global constraint satisfaction (see Figure~\ref{fig:qual_sudoku_hard}).

\paragraph{Robustness to $K$-Corrupted Sudoku.}
To test whether IPR can recover from more severe inconsistencies, we start from valid Sudoku grids and introduce controlled corruption by randomly swapping $K$ pairs of cells ($K \in \{1, 3, 5, 7, 9\}$), then allow all cells to be modified during refinement. As shown in Figure~\ref{fig:corrupt_sudoku}, for small $K$, validity is quickly restored within a few iterations, while for larger $K$, recovery is slower but continues to improve up to 50 iterations. This confirms that IPR can recover global consistency even from severely corrupted states, given sufficient iterations (see Figure~\ref{fig:qual_sudoku_corrupt}).

\subsection{Counting Polygons}

\textbf{Counting Polygons} requires generating a $128 \times 128$ image containing an FFHQ~\citep{karras2019style} face background, multiple polygons, and two digits. The digits specify the number of polygons and their vertex count, and the generated polygons must match these values exactly. This is an unconditional constrained generation task with no input conditions provided (see Appendix~\ref{app:counting_polygons} for details). We use SRMs trained on 256 patches of size $8 \times 8$.

\paragraph{Evaluation Metrics.}
We measure two metrics: \emph{Number Match Accuracy}, which checks whether the generated digits match the actual polygon count and vertex type, and \emph{Vertex Uniformity}, which checks whether all polygons in the image share the same number of vertices.

\begin{figure}[t]
    \centering
    \includegraphics[width=0.8\linewidth]{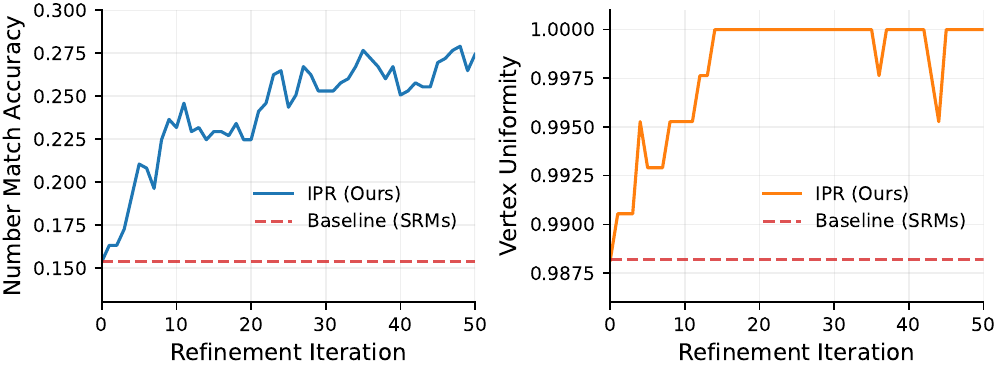}
    \caption{\textbf{IPR on Counting Polygons.} (Left) \emph{Number Match Accuracy} improves as iterations increase. (Right) \emph{Vertex Uniformity} reaches 100\% within 50 iterations.}
    \label{fig:counting_polygons}
\end{figure}

\paragraph{Results.}
As shown in Figure~\ref{fig:counting_polygons}, both metrics improve with IPR iterations. The baseline achieves 15.4\% on Number Match Accuracy and 98.8\% on Vertex Uniformity; after 50 iterations, these improve to 27.4\% (+12 pp) and 100\%, respectively. The model adjusts the digits to match the polygons, iteratively refining the context to satisfy the constraints. Qualitatively, we also observe that the FFHQ background and the polygon shapes become clearer alongside constraint satisfaction, suggesting that IPR improves overall image coherence beyond just the constrained elements (see Figure~\ref{fig:qual_counting}).

\subsection{Even Pixels}

\textbf{Even Pixels} requires generating a $32 \times 32$ image with exactly two colors, each occupying exactly half of the pixels. This is an unconditional constrained generation task that tests whether the model can satisfy a global counting constraint (see Appendix~\ref{app:even_pixels}). We use SRMs trained on patches of size $4 \times 4$.

\paragraph{Evaluation Metrics.}
We evaluate two aspects: \emph{Balance Accuracy} measures whether the two colors each occupy exactly 50\% of the pixels, and \emph{Saturation/Value Std} measures color consistency within each region---lower values indicate more uniform colors without noisy outliers.

\paragraph{Results.}
% As shown in Figure~\ref{fig:even_pixels}, Balance Accuracy improves by approximately 5\% after 50 IPR iterations. This gain alone may appear modest, but Balance Accuracy does not fully reflect generation quality---early iterations produce many pixels with noisy color values that happen to fall into the correct color class, masking visual artifacts.

% The Saturation/Value Std metrics better capture this effect. Over 50 iterations, Saturation Std drops from 0.0073 to 0.0027 (63\% reduction) and Value Std from 0.0047 to 0.0021 (55\% reduction), indicating that IPR substantially reduces color outliers and produces cleaner, more uniform regions. These results show that IPR improves overall image quality beyond just satisfying the counting constraint.

\begin{figure}[t]
    \centering
    \includegraphics[width=1.0\linewidth]{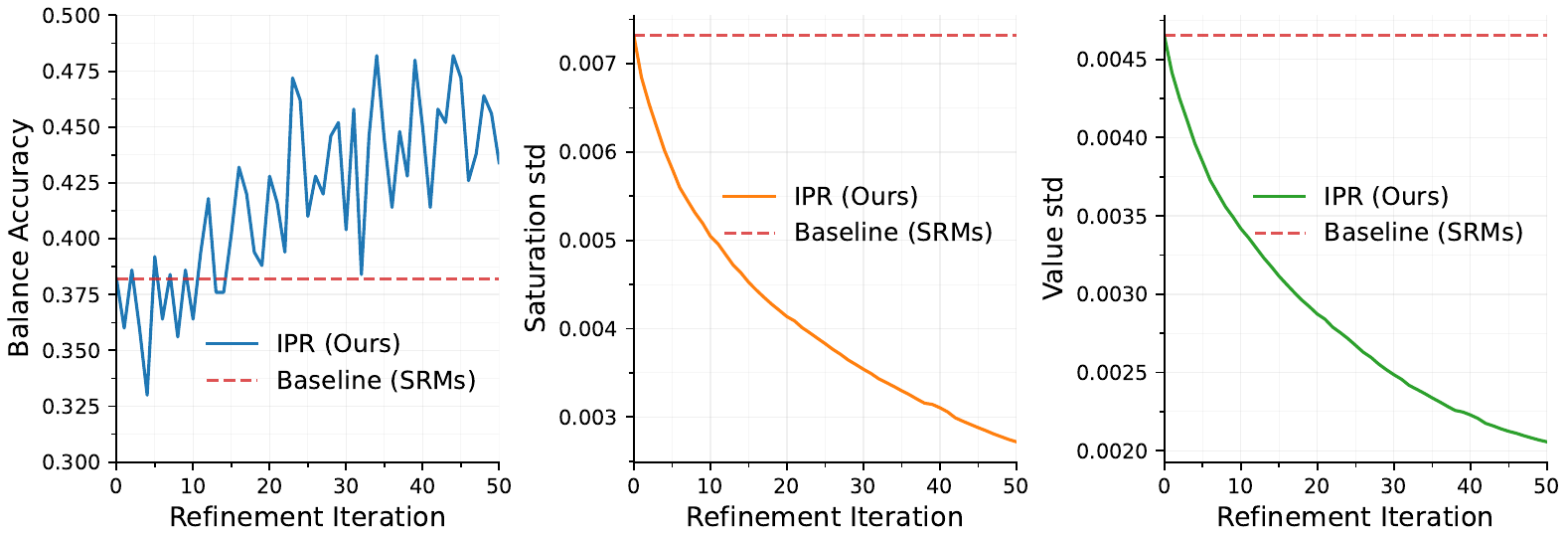}
    \caption{\textbf{IPR on Even Pixels.} Balance Accuracy (left) and color consistency measured by Saturation/Value std (middle, right) both improve with IPR iterations.}
    \label{fig:even_pixels}
\end{figure}

As shown in Figure~\ref{fig:even_pixels}, Balance Accuracy improves by approximately 5\% after 50 IPR iterations. Beyond the balance constraint itself, the color consistency within each region also improves significantly. The Saturation/Value Std metrics capture this effect. Over 50 iterations, Saturation Std drops from 0.0073 to 0.0027 (63\% reduction) and Value Std from 0.0047 to 0.0021 (55\% reduction), indicating that IPR produces cleaner, more uniform colors within each region. These results show that IPR improves both constraint satisfaction and overall image quality (see Figure~\ref{fig:qual_even}).

\subsection{Ablation Study}

\begin{figure}[t]
    \centering
    \begin{subfigure}[t]{0.48\textwidth}
        \centering
        \includegraphics[width=0.85\linewidth]{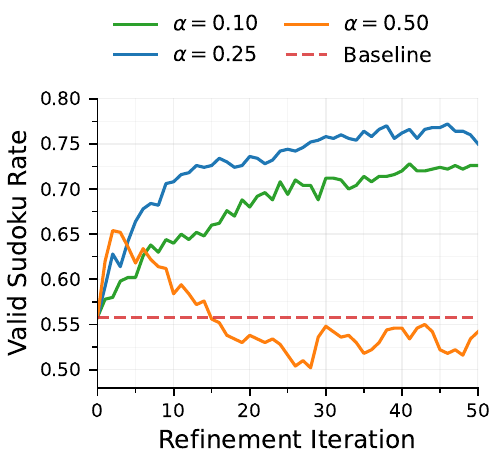}
        \caption{Effect of resampling ratio $\alpha$}
        \label{fig:ablation_ratio}
    \end{subfigure}
    \hfill
    \begin{subfigure}[t]{0.48\textwidth}
        \centering
        \includegraphics[width=0.85\linewidth]{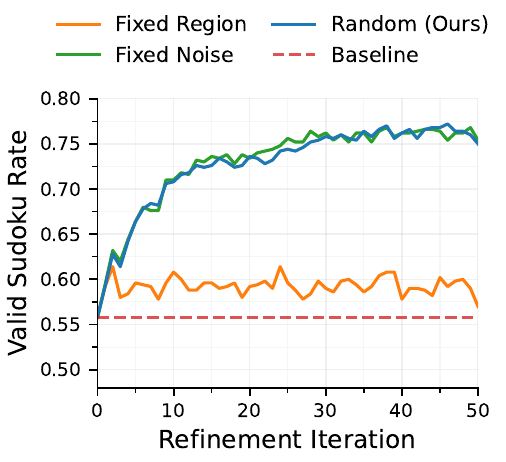}
        \caption{Effect of Fixed Noise and Region}
        \label{fig:ablation_noise}
    \end{subfigure}
    \caption{\textbf{Ablation studies on IPR hyperparameters.} Valid Sudoku rate (\%) on MNIST Sudoku HARD setting. (a) Resampling ratio $\alpha=0.25$ performs best. (b) Fixed Noise performs comparably to random noise, while Fixed Region selection degrades performance.}
    \label{fig:ablation}
\end{figure}

We conduct ablation studies on MNIST Sudoku (HARD setting) to analyze the key design choices of IPR. Results are shown in Figure~\ref{fig:ablation}.
\paragraph{Resampling Ratio.} 
The resampling ratio $\alpha$ controls the fraction of regions re-noised at each iteration. As shown in Figure~\ref{fig:ablation_ratio}, setting $\alpha = 0.25$ yields the best performance, improving the valid Sudoku rate from 55.8\% to 75.0\% after 50 iterations. A smaller ratio ($\alpha = 0.10$) leads to slower but steady improvement, reaching 72.6\%. Conversely, $\alpha = 0.50$ causes performance degradation to 54.2\%, falling below the baseline. This indicates that resampling too large a fraction discards valuable context along with errors, undermining the refinement process.

\paragraph{Fixed Noise vs. Random Noise.}
We examine whether IPR requires fresh noise sampling at each iteration or can function with fixed noise. Instead of sampling from $\mathcal{N}(0, I)$ at each re-noising step, we fix the noise to the original noise used during the initial SRMs generation. As shown in Figure~\ref{fig:ablation_noise}, both variants achieve comparable performance, indicating that the choice of noise has minimal impact on IPR. This suggests that the \emph{context} provided by already-generated regions, rather than the specific noise realization, is the primary factor driving refinement in sequential diffusion.

\paragraph{Fixed Region Selection.}
We further investigate whether the randomness in region selection is essential for IPR. Instead of randomly selecting regions to re-noise at each iteration, we fix the selection to the regions chosen in the first iteration. As shown in Figure~\ref{fig:ablation_noise}, this variant achieves only 57.0\% at 50 iterations, showing negligible improvement over the baseline (55.8\%). This confirms that varying the resampled regions across iterations is crucial for effective refinement: fixing the region selection prevents the model from propagating corrections across different regions, limiting the refinement to a static subset of the grid.

\section{Related Work}
\label{sec:related}

A growing line of work studies inference-time scaling for diffusion models
\citep{singhal2025general, li2024derivative, mctd, fast_mctd, lee2025adaptive, zhang2025inference, he2025scaling, zhang2025vfscale}.
For example, \citet{guo2025training} proposes a tree-search-based path steering method that branches into multiple candidates at each denoising step and selects among them using a value function, while \citet{lee2025adaptive} expands exploration by repeatedly selecting the top-$K$ particles and cycling them back to an intermediate noise level.
SMC/Feynman--Kac approaches similarly maintain multiple particles and perform reward-based reweighting and resampling at intermediate stages to bias sampling toward high-reward regions \citep{singhal2025general, kim2025test}.
Despite their differences, these methods commonly rely on an external reward model or verifier (or an objective-based scoring function) to rank, select, or preserve better samples, which limits applicability to tasks where such verifiers are unavailable or unreliable.

% In contrast, \citet{zhang2025vfscale} (VFScale) aims for verifier-free test-time scaling by using the diffusion model's intrinsic energy function as a verifier.
% It aligns the energy landscape with sample quality during training via MRNCL and KL regularization, and then increases test-time compute by coupling denoising with hMCTS to expand the number of sampled trajectories.
% However, VFScale ultimately scales inference through \emph{search and selection} driven by a scalar intrinsic score (energy).
% For problems with complex global constraints, it can be challenging for a single learned score to reliably reflect constraint satisfaction and long-range cross-region consistency, which can limit the practical reach of score-driven scaling.

In contrast, \citet{zhang2025vfscale} (VFScale) aims for verifier-free test-time scaling by using the diffusion model's intrinsic energy as an internal scoring signal.
It trains the energy to correlate with sample quality, then scales test-time compute by searching over denoising trajectories guided by this score.
However, VFScale ultimately relies on a single scalar score to drive search and selection.
For problems with complex global constraints, such a score may not reliably capture constraint satisfaction and long-range cross-region consistency, limiting the practical reach of score-driven scaling.

Unlike VFScale, \textbf{IPR} removes scoring and candidate selection altogether.
Instead, IPR scales compute by repeatedly \emph{editing} a single sample: it selectively re-noises a subset of regions and regenerates them conditioned on the remaining regions under the mixed-noise structure of sequential diffusion.
By keeping parts of the sample revisable after they are generated, IPR provides verifier- and reward-free inference-time scaling tailored to globally constrained generation in sequential diffusion.

\section{Discussion and Limitations}
\label{sec:discussion}

\paragraph{Computational cost.}
IPR increases inference cost linearly with the number of refinement iterations. With $\alpha{=}0.25$ and $R{=}50$, the total computation is roughly $13\times$ that of a single generation pass. This trade-off is inherent to inference-time scaling: improved quality comes at the expense of additional compute. In practice, the cost can be controlled by choosing smaller $R$ or $\alpha$ depending on the application's latency requirements.

\paragraph{Sensitivity to resampling ratio.}
The resampling ratio $\alpha$ must be carefully chosen. When $\alpha$ is too large ($\alpha{=}0.50$), performance degrades below the baseline because too much useful context is discarded along with errors. This suggests that IPR's effectiveness relies on preserving a sufficient amount of correct context at each iteration to guide regeneration.

\paragraph{Future directions.}
IPR currently selects regions uniformly at random. More informed strategies---such as targeting regions that violate constraints or learning a state-dependent selection policy---may improve efficiency. Adapting the resampling ratio or re-noising strength over iterations is another avenue for finer-grained refinement.

\section{Conclusion}
\label{sec:conclusion}

We introduced \textbf{Iterative Partial Refinement (IPR)}, an inference-time scaling method for sequential diffusion models trained on arbitrary noise levels. IPR repeatedly selects a subset of regions, re-noises only the selected regions, and regenerates them conditioned on the remaining regions. This keeps parts of the sample revisable during sampling, providing a simple way to correct inconsistencies that can persist under the default one-pass generation trajectory. IPR requires no additional training, no guidance, and no external verifier; it leverages the model's mixed-noise denoising capability to revise a generated sample and improve global consistency at inference time.

Across three constrained generation benchmarks, IPR consistently improves constraint satisfaction---raising the valid Sudoku rate from 55.8\% to over 75\%, Number Match Accuracy from 15.4\% to 27.4\%, and reducing color outliers in Even Pixels---while also improving overall visual quality. By randomly varying which regions are resampled, IPR propagates contextual improvements across the entire sample over iterations, rather than refining only a fixed subset. The resampling ratio is critical to balance preserving useful context with allowing meaningful revision.

\subsubsection*{Acknowledgments}
This work was supported by Basic Research Laboratory Program (No. RS-2024-00414822) through the National Research Foundation of Korea (NRF) grant funded by the Korea government (MSIT). This work was also supported by Institute of Information \& communications Technology Planning \& Evaluation (IITP) grant funded by the Korea government (MSIT) (No. RS-2024-00509279, Global AI Frontier Lab). We thank the members of the Machine Learning and Mind Lab (MLML) for their valuable discussions and assistance.

\newpage

\bibliography{iclr2026_conference}

@article{ho2020denoising,
  title   = {Denoising diffusion probabilistic models},
  author  = {Ho, Jonathan and Jain, Ajay and Abbeel, Pieter},
  journal = {Advances in neural information processing systems},
  volume  = {33},
  pages   = {6840--6851},
  year    = {2020}
}

@article{song2020denoising,
  title   = {Denoising diffusion implicit models},
  author  = {Song, Jiaming and Meng, Chenlin and Ermon, Stefano},
  journal = {arXiv preprint arXiv:2010.02502},
  year    = {2020}
}

@article{lipman2022flow,
  title   = {Flow matching for generative modeling},
  author  = {Lipman, Yaron and Chen, Ricky TQ and Ben-Hamu, Heli and Nickel, Maximilian and Le, Matt},
  journal = {arXiv preprint arXiv:2210.02747},
  year    = {2022}
}

@article{chen2024diffusion,
  title   = {Diffusion forcing: Next-token prediction meets full-sequence diffusion},
  author  = {Chen, Boyuan and Mart{\'\i} Mons{\'o}, Diego and Du, Yilun and Simchowitz, Max and Tedrake, Russ and Sitzmann, Vincent},
  journal = {Advances in Neural Information Processing Systems},
  volume  = {37},
  pages   = {24081--24125},
  year    = {2024}
}

@article{wewer2025spatial,
  title   = {Spatial reasoning with denoising models},
  author  = {Wewer, Christopher and Pogodzinski, Bart and Schiele, Bernt and Lenssen, Jan Eric},
  journal = {arXiv preprint arXiv:2502.21075},
  year    = {2025}
}

@article{wu2023ar,
  title   = {Ar-diffusion: Auto-regressive diffusion model for text generation},
  author  = {Wu, Tong and Fan, Zhihao and Liu, Xiao and Zheng, Hai-Tao and Gong, Yeyun and Jiao, Jian and Li, Juntao and Guo, Jian and Duan, Nan and Chen, Weizhu and others},
  journal = {Advances in Neural Information Processing Systems},
  volume  = {36},
  pages   = {39957--39974},
  year    = {2023}
}

@inproceedings{zhang2024tedi,
  title     = {Tedi: Temporally-entangled diffusion for long-term motion synthesis},
  author    = {Zhang, Zihan and Liu, Richard and Hanocka, Rana and Aberman, Kfir},
  booktitle = {ACM SIGGRAPH 2024 Conference Papers},
  pages     = {1--11},
  year      = {2024}
}

@article{li2024autoregressive,
  title   = {Autoregressive image generation without vector quantization},
  author  = {Li, Tianhong and Tian, Yonglong and Li, He and Deng, Mingyang and He, Kaiming},
  journal = {Advances in Neural Information Processing Systems},
  volume  = {37},
  pages   = {56424--56445},
  year    = {2024}
}

@article{singhal2025general,
  title   = {A general framework for inference-time scaling and steering of diffusion models},
  author  = {Singhal, Raghav and Horvitz, Zachary and Teehan, Ryan and Ren, Mengye and Yu, Zhou and McKeown, Kathleen and Ranganath, Rajesh},
  journal = {arXiv preprint arXiv:2501.06848},
  year    = {2025}
}

@article{li2024derivative,
  title   = {Derivative-free guidance in continuous and discrete diffusion models with soft value-based decoding},
  author  = {Li, Xiner and Zhao, Yulai and Wang, Chenyu and Scalia, Gabriele and Eraslan, Gokcen and Nair, Surag and Biancalani, Tommaso and Ji, Shuiwang and Regev, Aviv and Levine, Sergey and others},
  journal = {arXiv preprint arXiv:2408.08252},
  year    = {2024}
}

@article{kim2025test,
  title   = {Test-time alignment of diffusion models without reward over-optimization},
  author  = {Kim, Sunwoo and Kim, Minkyu and Park, Dongmin},
  journal = {arXiv preprint arXiv:2501.05803},
  year    = {2025}
}

@article{guo2025training,
  title   = {Training-free guidance beyond differentiability: Scalable path steering with tree search in diffusion and flow models},
  author  = {Guo, Yingqing and Yang, Yukang and Yuan, Hui and Wang, Mengdi},
  journal = {arXiv preprint arXiv:2502.11420},
  year    = {2025}
}

@inproceedings{lee2025adaptive,
  title     = {Adaptive Inference-Time Scaling via Cyclic Diffusion Search},
  author    = {Lee, Gyubin and Truong, Bao N Nguyen and Yoon, Jaesik and Lee, Dongwoo and Kim, Minsu and Bengio, Yoshua and Ahn, Sungjin},
  booktitle = {The Thirty-ninth Annual Conference on Neural Information Processing Systems},
  year      = {2025}
}

@article{zhang2025inference,
  title   = {Inference-time scaling of diffusion models through classical search},
  author  = {Zhang, Xiangcheng and Lin, Haowei and Ye, Haotian and Zou, James and Ma, Jianzhu and Liang, Yitao and Du, Yilun},
  journal = {arXiv preprint arXiv:2505.23614},
  year    = {2025}
}

@article{he2025scaling,
  title   = {Scaling Image and Video Generation via Test-Time Evolutionary Search},
  author  = {He, Haoran and Liang, Jiajun and Wang, Xintao and Wan, Pengfei and Zhang, Di and Gai, Kun and Pan, Ling},
  journal = {arXiv preprint arXiv:2505.17618},
  year    = {2025}
}

@article{zhang2025vfscale,
  title   = {VFScale: Intrinsic Reasoning through Verifier-Free Test-time Scalable Diffusion Model},
  author  = {Zhang, Tao and Pan, Jia-Shu and Feng, Ruiqi and Wu, Tailin},
  journal = {arXiv preprint arXiv:2502.01989},
  year    = {2025}
}

@inproceedings{karras2019style,
  title     = {A style-based generator architecture for generative adversarial networks},
  author    = {Karras, Tero and Laine, Samuli and Aila, Timo},
  booktitle = {Proceedings of the IEEE/CVF conference on computer vision and pattern recognition},
  pages     = {4401--4410},
  year      = {2019}
}

@inproceedings{mctd,
  title={Monte Carlo Tree Diffusion for System 2 Planning},
  author={Yoon, Jaesik and Cho, Hyeonseo and Baek, Doojin and Bengio, Yoshua and Ahn, Sungjin},
  booktitle={Forty-second International Conference on Machine Learning},
  year={2025}
}

@article{fast_mctd,
  title={Fast Monte Carlo Tree Diffusion: 100x Speedup via Parallel Sparse Planning},
  author={Yoon, Jaesik and Cho, Hyeonseo and Bengio, Yoshua and Ahn, Sungjin},
  journal={arXiv preprint arXiv:2506.09498},
  year={2025}
}
\bibliographystyle{iclr2026_conference}

\newpage
\appendix
\section{Use of Large Language Models}
We used large language models only for grammar checking and sentence correction.

\section{Qualitative Results}
\label{qual_appendix}
\begin{figure}[h]
    \centering
    \includegraphics[width=\linewidth]{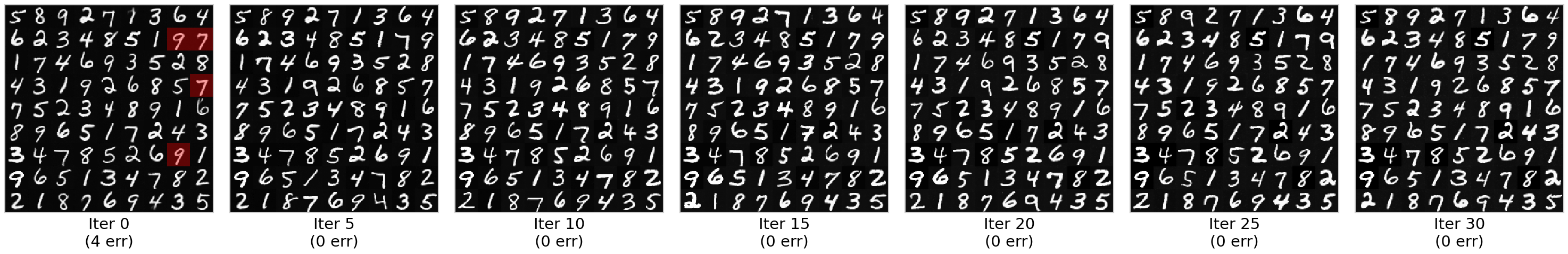}
    \includegraphics[width=\linewidth]{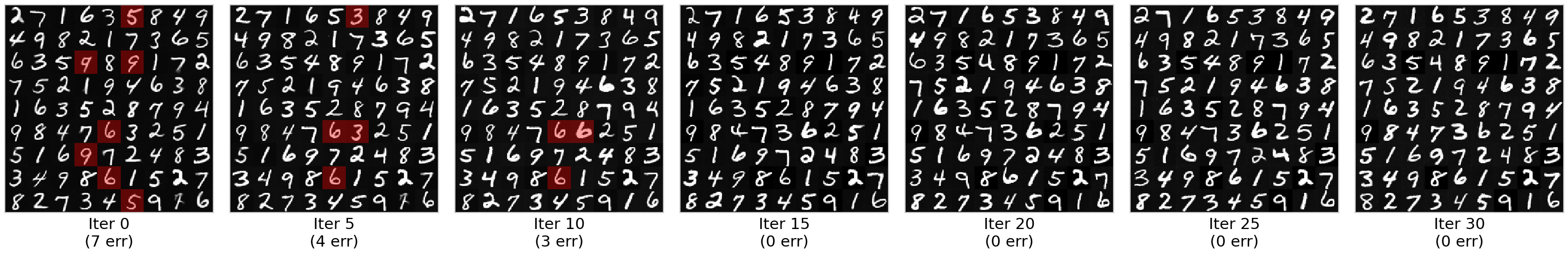}
    \includegraphics[width=\linewidth]{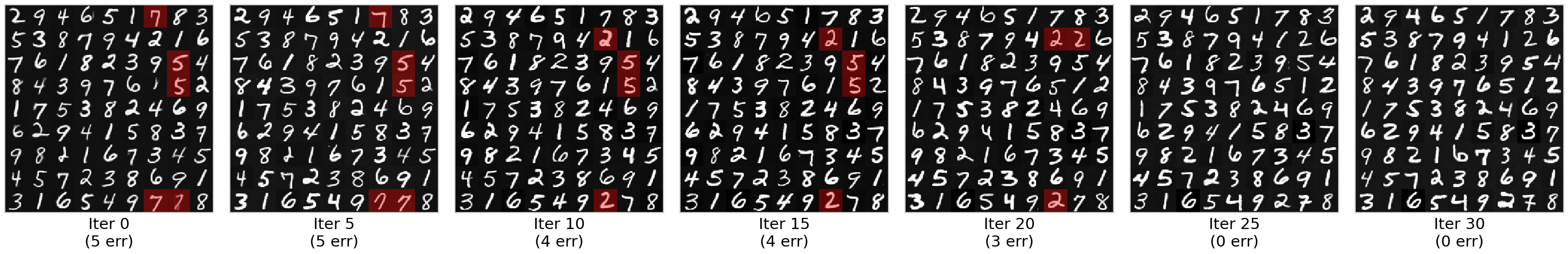}
    \caption{\textbf{Generated examples on MNIST Sudoku (HARD).} Each column shows the generated image at a different IPR iteration, progressing from left to right. Red backgrounds indicate cells that violate Sudoku constraints; violations decrease as refinement progresses.}
    \label{fig:qual_sudoku_hard}
\end{figure}

\begin{figure}[h]
    \centering
    \includegraphics[width=\linewidth]{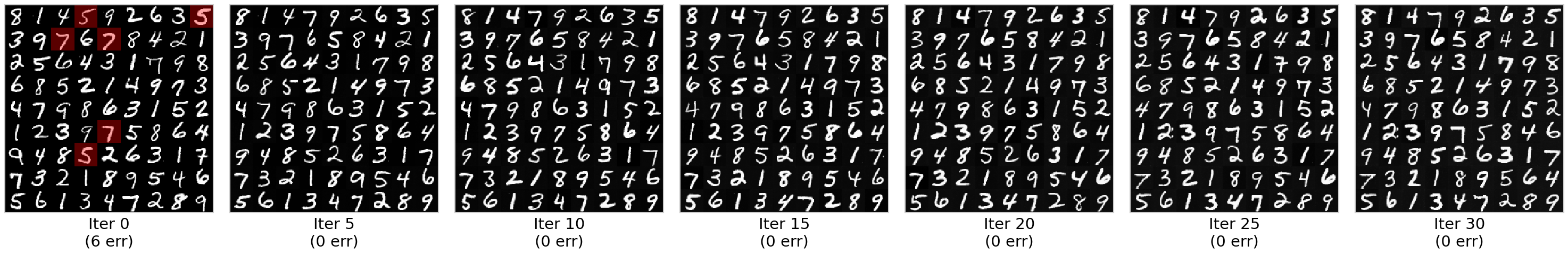}
    \includegraphics[width=\linewidth]{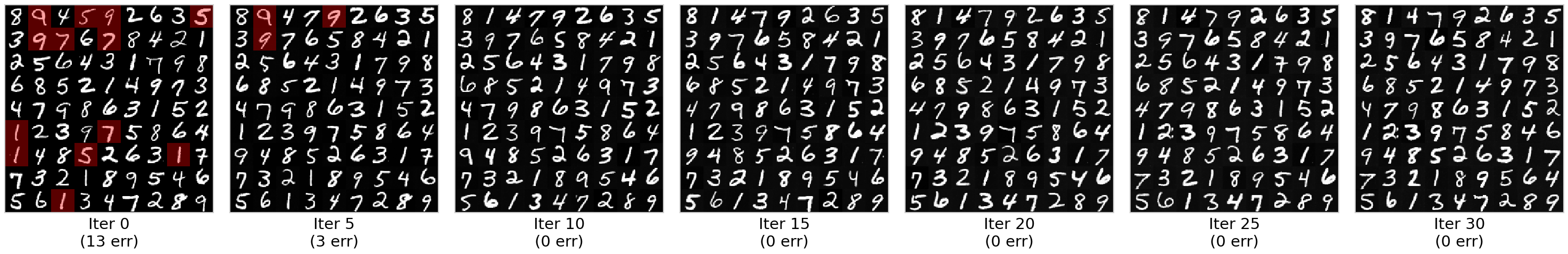}
    \includegraphics[width=\linewidth]{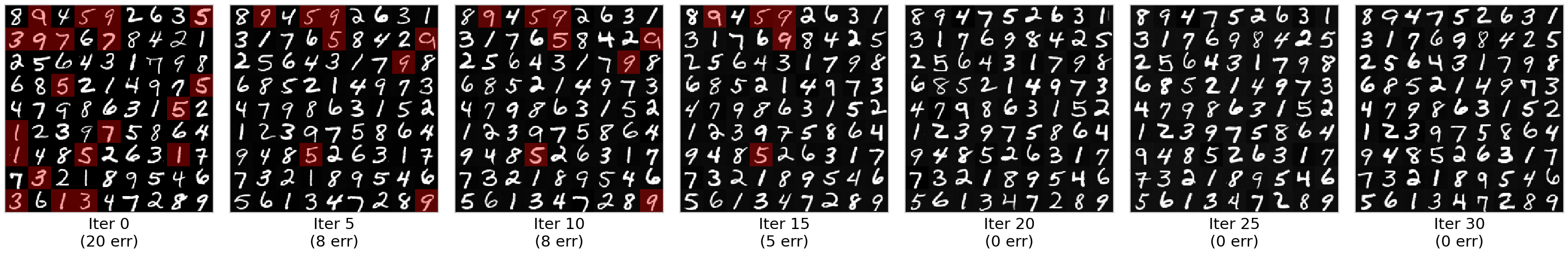}
    \caption{\textbf{Generated examples on $K$-Corrupted Sudoku ($K \in \{1, 3, 5\}$).} Each row corresponds to $K=1, 3, 5$ from top to bottom. Each column shows the generated image at a different IPR iteration, progressing from left to right. Red backgrounds indicate cells that violate Sudoku constraints; violations decrease as refinement progresses.}
    \label{fig:qual_sudoku_corrupt}
\end{figure}

\begin{figure}[h]
    \centering
    \includegraphics[width=\linewidth]{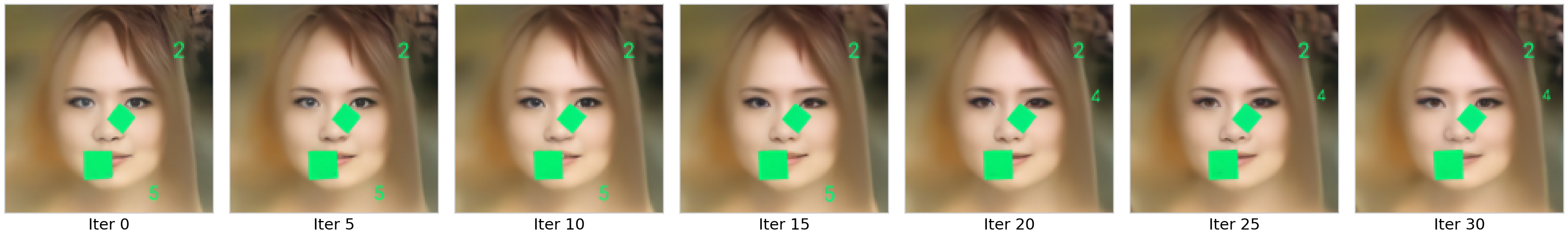}
    \includegraphics[width=\linewidth]{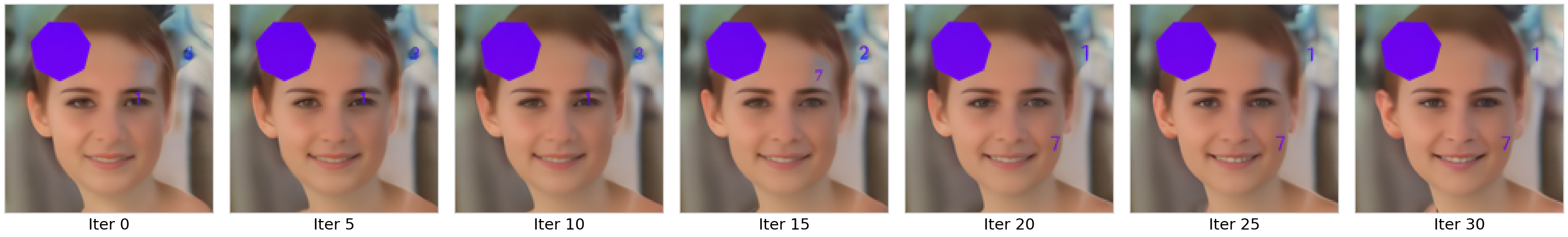}
    \includegraphics[width=\linewidth]{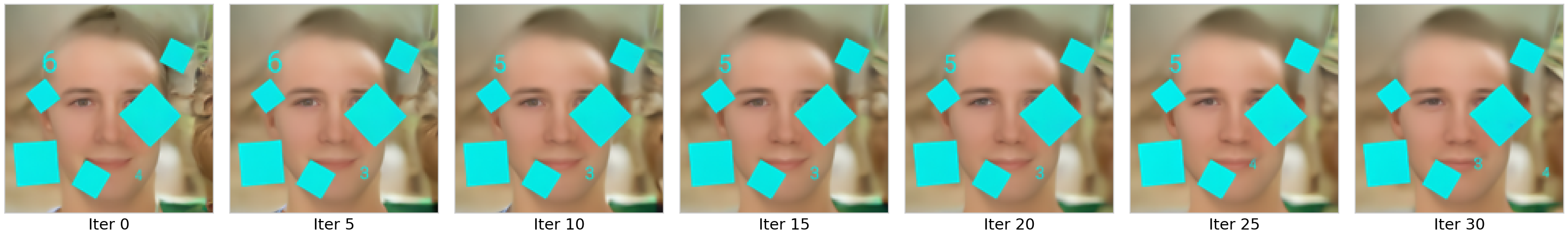}
    \caption{\textbf{Generated examples on Counting Polygons.} Each column shows the generated image at a different IPR iteration, progressing from left to right. The digit and polygon count converge to match the constraint.}
    \label{fig:qual_counting}
\end{figure}

\begin{figure}[h]
    \centering
    \includegraphics[width=\linewidth]{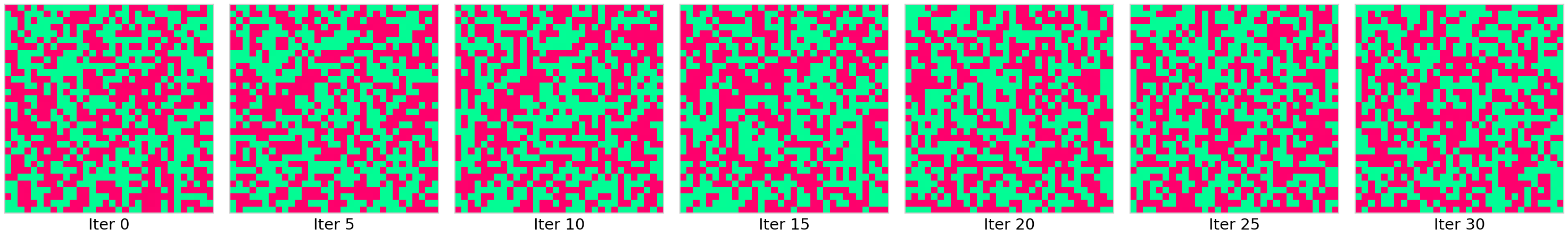}
    \caption{\textbf{Generated examples on Even Pixels.} Each column shows the generated image at a different IPR iteration, progressing from left to right. Color regions become more uniform and balanced across iterations.}
    \label{fig:qual_even}
\end{figure}

\section{Experiment Details}

\subsection{Spatial Reasoning Models}
We use SRMs~\citep{wewer2025spatial} as our sequential diffusion backbone for inference.

We use the pretrained models provided by the authors. Checkpoints were downloaded from \url{https://github.com/Chrixtar/SRM/releases}. The specific checkpoints used are:
\begin{itemize}
    \item \textbf{MNIST Sudoku}: \texttt{ms1000\_28/paper/checkpoints/last.ckpt}
    \item \textbf{Counting Polygons}: \texttt{cp\_ffhq\_8/paper/checkpoints/last.ckpt}
    \item \textbf{Even Pixels}: \texttt{ep\_4/paper/checkpoints/last.ckpt}
\end{itemize}

\subsection{MNIST Sudoku}
\label{app:mnist_sudoku}

\subsubsection{Task Description}
MNIST Sudoku is a constrained generation benchmark proposed by \citet{wewer2025spatial} that combines the visual complexity of handwritten digit generation with the combinatorial constraints of Sudoku. Each sample is a $9 \times 9$ grid where every cell contains a $28 \times 28$ MNIST digit image. The model must generate digits that are both visually realistic and satisfy standard Sudoku rules: each row, column, and $3 \times 3$ sub-grid must contain the digits 1--9 without repetition.

\subsubsection{Experimental Setting}
The benchmark supports varying difficulty by controlling the number of pre-filled hint cells $N_{\text{hint}}$. We adopt the \textbf{HARD} setting from \citet{wewer2025spatial}, where $N_{\text{hint}}$ is sampled uniformly from $\{0, \dots, 26\}$ for each sample. The positions of the hint cells are selected uniformly at random from the 81 cells in the grid, and the remaining $81 - N_{\text{hint}}$ cells are masked and must be generated by the model. This setting covers a wide range of difficulty, from nearly empty grids (purely generative) to moderately constrained puzzles (completion). During IPR, hint cells are fixed and excluded from resampling.

\subsubsection{Evaluation Methodology}
Since the output consists of images, we first map each generated cell to a digit class $d \in \{1, \dots, 9\}$ using a pre-trained CNN classifier provided by the benchmark~\citep{wewer2025spatial}. Let $G$ denote the resulting $9 \times 9$ symbolic grid. A generated sample is considered valid if and only if $G$ contains no duplicate digits in any row, column, or sub-grid. We report the \textbf{Sudoku Validity Rate}, i.e., the percentage of valid solutions over the test set.

\subsection{Counting Polygons}
\label{app:counting_polygons}

\subsubsection{Task Description}
Counting Polygons is an unconditional constrained generation benchmark proposed by \citet{wewer2025spatial}. The model must generate a $128 \times 128$ image containing an FFHQ face background, multiple polygons, and two digits. The two digits specify the number of polygons and their vertex count, respectively, and the generated polygons must match these values exactly. Crucially, no input conditions are provided: the model must jointly generate all components---background, polygons, and digits---such that they are internally consistent.

\subsubsection{Training Dataset Construction}
The training dataset is constructed by \citet{wewer2025spatial} as follows. Each sample is produced by overlaying randomly generated geometric polygons onto face images from the FFHQ dataset~\citep{karras2019style}, which serve as diverse and realistic backgrounds:
\begin{itemize}
    \item \textbf{Background}: A $128 \times 128$ face image is sampled from the FFHQ dataset.
    \item \textbf{Object Sampling}: The number of polygons ($N_{\text{poly}}$) and the number of vertices per polygon ($N_{\text{vert}}$) are sampled uniformly from predefined ranges (e.g., $N_{\text{vert}} \in [3, 7]$).
    \item \textbf{Appearance \& Placement}: Polygons are drawn with random positions, scales, and rotations. To ensure visibility against the complex background, the fill color is dynamically selected by computing the HSV color histogram of the background image and choosing the hue with the minimum frequency, thereby maximizing color contrast.
\end{itemize}

\subsubsection{Evaluation Methodology}
Since this is an unconditional generation task, evaluation focuses on the internal consistency of the generated image. We employ a fine-tuned ResNet-50 classifier, provided by the benchmark~\citep{wewer2025spatial}, to predict three attributes from the generated image: polygon count, vertex count, and uniformity. We report two metrics: \emph{Number Match Accuracy}, which checks whether the generated digits match the actual polygon count and vertex type, and \emph{Vertex Uniformity}, which checks whether all polygons in the image share the same number of vertices.

\subsection{Even Pixels}
\label{app:even_pixels}

\subsubsection{Task Description}
Even Pixels is a constrained generation benchmark designed to test the model's ability to satisfy precise constraints on pixel-level statistics. Each sample is a $32 \times 32$ image composed of exactly two distinct colors. The global constraint is that each color must occupy exactly 50\% of the pixels.

\subsubsection{Dataset Construction}
The Even Pixels dataset~\citep{wewer2025spatial} is constructed to isolate the challenge of generating a balanced hue distribution while maintaining constant saturation and value.
\begin{itemize}
    \item \textbf{Color Selection}: Images are generated in the HSV color space. A base hue $h_1$ is sampled uniformly from $[0, 0.5)$. The second hue is set to $h_2 = h_1 + 0.5$, ensuring the two colors are separated by $180^\circ$ in the color wheel to maximize contrast. Saturation ($S$) and Value ($V$) are fixed to constant values (e.g., $S=1.0, V=0.7$) across the entire image.
    \item \textbf{Pixel Assignment}: A random binary mask $M \in \{0, 1\}^{32 \times 32}$ is generated such that $\sum_{i,j} M_{i,j} = 512$. Pixels corresponding to 0 are assigned $h_1$, and those corresponding to 1 are assigned $h_2$.
\end{itemize}

\subsubsection{Evaluation Methodology}
Evaluating the generated images requires verifying whether the two dominant colors appear in exactly equal proportions. The benchmark employs a histogram-based clustering approach to robustly count pixels for each color without assuming predefined color values.

\paragraph{Procedure.}
\begin{enumerate}
    \item \textbf{Color Space Conversion}: The generated RGB image is converted to the HSV color space. The analysis focuses on the Hue ($H$) channel.
    \item \textbf{Histogram Analysis}: A histogram of the Hue channel (256 bins) is computed, and the two most prominent peaks, $p_1$ and $p_2$, are identified.
    \item \textbf{Dynamic Thresholding}: Decision boundaries $b_1, b_2$ between the two colors are determined by finding the midpoints between the peaks in the circular Hue space:
    \begin{equation}
        b_1 = \frac{p_1 + p_2}{2}, \quad b_2 = \frac{p_1 + p_2 + 256}{2} \pmod{256}.
    \end{equation}
    \item \textbf{Pixel Counting}: All pixels with hue values falling between $b_1$ and $b_2$ are assigned to one cluster. Let $N_{c1}$ be the pixel count of this cluster.
\end{enumerate}

\paragraph{Metrics.}
\textbf{Balance Accuracy} is defined as the percentage of images where the pixel count error $|N_{c1} - 512|$ is exactly 0. Additionally, \textbf{Saturation/Value Std} is measured to assess internal color consistency within each region.

\section{Inference Hyperparameters}
\label{app:hyperparameters}

We summarize the inference-time hyperparameters used in our experiments.

\subsection{Parameter Definitions}
\begin{itemize}
    \item \texttt{init\_overlap\_ratio}: Controls the degree of scheduling overlap between regions during the initial generation (SRM). Larger values allow more concurrent denoising.
    \item \texttt{init\_steps\_per\_patch}: The number of denoising steps allocated to each region during the initial generation.
    \item \texttt{ipr\_overlap\_ratio}: The scheduling overlap ratio used during IPR refinement.
    \item \texttt{ipr\_steps\_per\_patch}: The number of denoising steps applied to each re-sampled region during IPR refinement.
    \item \texttt{stochasticity}: Controls the amount of randomness injected during diffusion sampling (for both initial generation and refinement).
    \item \texttt{resampling\_ratio}: Controls the fraction of regions re-noised at each IPR iteration.
    \item \texttt{iteration}: The total number of IPR refinement iterations performed.
\end{itemize}

\subsection{Task-Specific Settings}
For all experiments, we fixed the following hyperparameters for each task and varied \texttt{resampling\_ratio} and \texttt{iteration} to analyze their effects.

\begin{table}[h]
    \centering
    \caption{Fixed inference hyperparameters for each task.}
    \label{tab:hyperparameters}
    \begin{tabular}{lccc}
        \toprule
        \textbf{Hyperparameter} & \textbf{MNIST Sudoku} & \textbf{Counting Polygons} & \textbf{Even Pixels} \\
        \midrule
        \texttt{init\_overlap\_ratio} & 0.0 & 0.9 & 0.9 \\
        \texttt{init\_steps\_per\_patch} & 3 & 10 & 30 \\
        \texttt{ipr\_overlap\_ratio} & 0.8 & 0.9 & 0.9 \\
        \texttt{ipr\_steps\_per\_patch} & 10 & 10 & 30 \\
        \texttt{stochasticity} & 0.5 & 0.5 & 0.5 \\
        \bottomrule
    \end{tabular}
\end{table}

\end{document}